\documentclass[runningheads]{llncs}
\usepackage{graphicx}
\usepackage[acronym]{glossaries}
\usepackage{placeins}
\usepackage[misc]{ifsym}
\usepackage{xcolor}
\usepackage{url}
\makeglossaries{}
\newacronym{cnn}{CNN}{Convolutional Neural Network}
\newacronym{cbir}{CBIR}{Content-based Image Retrieval}
\newacronym{ed}{ED}{Euclidean Distance}
\newacronym{bc}{BC}{Bhattacharya Coefficient}
\newacronym{md}{MD}{Mahalanobis Distance}
\begin{document}
%
\title{Integrating Visual and Semantic Similarity Using Hierarchies for Image Retrieval}
\titlerunning{Visual and Semantic Similarity Using Hierarchies for CBIR}
%

\author{Aishwarya Venkataramanan\inst{1,2,3}\orcidID{0000-0002-6100-0034}[\Letter] \and 
Martin Laviale\inst{1,3}\orcidID{0000-0002-9719-7158} \and
C\'edric Pradalier\inst{2,3}\orcidID{0000-0002-1746-2733}}

\institute{Laboratoire Interdisciplinaire des Environnements Continentaux, Universit\'e de Lorraine, Metz, France 
\email{aishwarya.venkataramanan@univ-lorraine.fr}
 \and Georgia Tech Europe- International Research Lab Georgia Tech - CNRS IRL 2958, Metz, France
 \and LTER- ``Zone Atelier Moselle", France
}

%
\authorrunning{A. Venkataramanan et al.}
%
%
\maketitle              
\begin{abstract}
Most of the research in content-based image retrieval (CBIR) focus on developing robust feature representations that can effectively retrieve instances from a database of images that are visually similar to a query. However, the retrieved images sometimes contain results that are not semantically related to the query. To address this, we propose a method for CBIR that captures both visual and semantic similarity using a visual hierarchy. The hierarchy is constructed by merging classes with overlapping features in the latent space of a deep neural network trained for classification, assuming that overlapping classes share high visual and semantic similarities. Finally, the constructed hierarchy is integrated into the distance calculation metric for similarity search. Experiments on standard datasets: CUB-200-2011 and CIFAR100, and a real-life use case using diatom microscopy images show that our method achieves superior performance compared to the existing methods on image retrieval.

\end{abstract}

\section{Introduction}

In recent years, the exponential growth of image data has made effective management of such data increasingly important. One way to manage this data is through \gls{cbir}, where a query image is used to retrieve images from the database that are visually similar to it~\cite{chen2022deep}. The comparison is performed on the features extracted from the images, represented by a set of numerical descriptors or vectors. The search is performed on these descriptors using a similarity measure such as the Euclidean or cosine distance.
The challenge is to extract robust feature representations, such that objects belonging to the same category are close by, despite the changes in view-point, background and illumination. Recent methods employing \gls{cnn} for feature extraction have proven to be robust and effective over the classical hand-crafted feature extractors~\cite{wan2014deep}.
A commonly adopted approach is to extract feature vectors from the latent space of a trained \gls{cnn} classifier. Deep classifiers use cross-entropy loss that learns feature representations to group together objects belonging to the same category, while separating them from the rest. Thus, given a query image, the retrieved data from the local neighbourhood in the latent space should ideally consist of images belonging to the same classification category. 

\begin{figure*}[t]
    \centering
    \begin{tabular}{c|c@{\hspace{0.05cm}}c@{\hspace{0.05cm}}c@{\hspace{0.05cm}}c@{\hspace{0.05cm}}c|ccccc}
    \multicolumn{1}{c}{Query} & \multicolumn{5}{c}{Retrieved images - Cosine Distance} & \multicolumn{5}{c}{Retrieved images - Ours}\\ 
    \vspace{-0.7em}
    \includegraphics[width=0.08\linewidth]{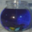} & 
    \includegraphics[width=0.08\linewidth]{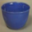} & \includegraphics[width=0.08\linewidth]{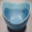} & \includegraphics[width=0.08\linewidth]{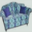} & \includegraphics[width=0.08\linewidth]{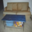} & \includegraphics[width=0.08\linewidth]{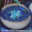} &   
    \includegraphics[width=0.08\linewidth]{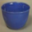} & \includegraphics[width=0.08\linewidth]{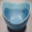} & \includegraphics[width=0.08\linewidth]{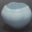} & \includegraphics[width=0.08\linewidth]{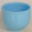} & \includegraphics[width=0.08\linewidth]{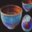} \\
    \tiny Bowl & \tiny Bowl & \tiny Bowl & \tiny Couch & \tiny Couch & \tiny Bowl & \tiny Bowl & \tiny Bowl & \tiny Bowl & \tiny Bowl & \tiny Bowl \\
    \vspace{-0.7em}
    \includegraphics[width=0.08\linewidth]{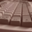} & 
    \includegraphics[width=0.08\linewidth]{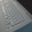} & \includegraphics[width=0.08\linewidth]{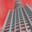} & \includegraphics[width=0.08\linewidth]{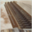} & \includegraphics[width=0.08\linewidth]{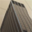} & \includegraphics[width=0.08\linewidth]{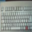} &     \includegraphics[width=0.08\linewidth]{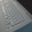} & \includegraphics[width=0.08\linewidth]{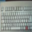} & \includegraphics[width=0.08\linewidth]{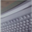} & \includegraphics[width=0.08\linewidth]{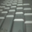} & \includegraphics[width=0.08\linewidth]{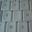} \\
    \tiny Keyboard & \tiny Keyboard & \tiny Skyscraper & \tiny Skyscraper & \tiny Skyscraper & \tiny Keyboard & \tiny Keyboard & \tiny Keyboard & \tiny Keyboard & \tiny Keyboard & \tiny Keyboard \\
    \vspace{-0.7em}
    \includegraphics[width=0.08\linewidth]{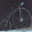} & 
    \includegraphics[width=0.08\linewidth]{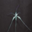} & \includegraphics[width=0.08\linewidth]{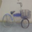} & \includegraphics[width=0.08\linewidth]{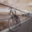} & \includegraphics[width=0.08\linewidth]{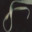} & \includegraphics[width=0.08\linewidth]{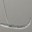} &     \includegraphics[width=0.08\linewidth]{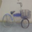} & \includegraphics[width=0.08\linewidth]{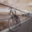} & \includegraphics[width=0.08\linewidth]{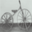} & \includegraphics[width=0.08\linewidth]{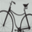} & \includegraphics[width=0.08\linewidth]{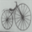} \\
    \tiny Bicycle & \tiny Spider & \tiny Bicycle & \tiny Bicycle & \tiny Worm & \tiny Worm & \tiny Bicycle & \tiny Bicycle & \tiny Bicycle & \tiny Bicycle & \tiny Bicycle \\
    \vspace{-0.7em}
    \includegraphics[width=0.08\linewidth]{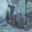} & 
    \includegraphics[width=0.08\linewidth]{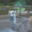} & \includegraphics[width=0.08\linewidth]{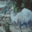} & \includegraphics[width=0.08\linewidth]{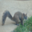} & \includegraphics[width=0.08\linewidth]{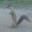} & \includegraphics[width=0.08\linewidth]{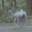} &     \includegraphics[width=0.08\linewidth]{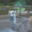} & \includegraphics[width=0.08\linewidth]{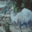} & \includegraphics[width=0.08\linewidth]{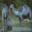} & \includegraphics[width=0.08\linewidth]{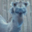} & \includegraphics[width=0.08\linewidth]{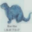} \\
    \tiny Camel & \tiny Camel & \tiny Camel & \tiny Squirrel & \tiny Squirrel & \tiny Camel & \tiny Camel & \tiny Camel & \tiny Camel & \tiny Camel & \tiny Otter \\
    \end{tabular}
    \caption{Examples of top-5 images retrieved using cosine distance and our method on CIFAR100. Our method incorporates hierarchy while ranking image similarity, with retrieves images that are more visually and semantically relevant. }
    \label{fig:retrievals}
\end{figure*}

Despite the ability of the \gls{cnn}s to learn discriminative representations, it has been observed that sometimes, images unrelated to the query are retrieved~\cite{barz2019hierarchy,park2002ranking}. These images are typically visually similar to the query, but semantically very different. Fig.~\ref{fig:retrievals} illustrates this for the images retrieved using cosine distance on CIFAR100~\cite{krizhevsky2009learning}. While the retrieved images have certain visual similarities with the query, some of them are semantically very different. 
One way to address this is by incorporating domain knowledge while training, so that the network learns feature representations that are both visually and semantically relevant~\cite{barz2019hierarchy}. The semantic data is obtained 
from expert domain knowledge or lexical databases such as WordNet~\cite{miller1995wordnet}. However, incorporating external domain knowledge during training can be challenging since it would require significant changes in the training procedure, and can be detrimental when there are inconsistencies between visual and semantic similarities~\cite{brust2019not}. Also, it assumes that the specialized domain knowledge is available, which can be hard to obtain or irrelevant in certain situations, such as, querying the images of people or vehicles. In this paper, we demonstrate that the visual hierarchies obtained from the representations learned by the classification network are rich in visual and semantic information, 
achieving superior retrieval performance.
Contrary to other approaches, our method does not require specialized domain knowledge or modifications to the training procedure, making it easy to integrate into an off-the-shelf classification network. 

Our method relies on two key aspects for the retrieval: construction of a hierarchy that is visually and semantically meaningful, and integration of hierarchy into the distance calculation metric for similarity search. 
Our method for hierarchy construction involves identifying the classes in the latent space whose feature representations overlap with one another. This is based on the assumption that the overlapping classes are those that are very similar to each other, both visually and semantically. The clusters of overlapping classes are then grouped together into the same class at a higher hierarchy level. This process is repeated at different levels of the hierarchy until there are no overlaps in the feature representations, resulting in a hierarchical structure that captures both visual and semantic similarities between images. Once the hierarchy is obtained, we combine the cosine distance and a hierarchy-based distance to obtain a metric for ranking the images for retrieval. 
Experimental results show that our method improves image retrieval performance on two standard benchmarks: CIFAR100~\cite{krizhevsky2009learning} and CUB-200-2011~\cite{wah2011caltech}, and a real life diatom image retrieval use case. Additionally, we demonstrate the robustness of our method by evaluating on images subject to perturbations that mimic practical difficulties encountered during acquisition.

Our contributions are as follows: (i) we develop a method to construct a hierarchy that is visually and semantically relevant and show that incorporating this hierarchy in the distance metric enhances the image retrieval performance, (ii) we design the method such that it does not require any additional domain knowledge and introduces no modifications to the model architecture or training procedure. The code is available in \url{https://github.com/vaishwarya96/Hierarchy-image-retrieval.git}


\section{Related Works}


The role of visual attributes and semantics is crucial in describing images for identification or retrieval. Hierarchies are an intuitive way to express this, and several methods have leveraged it for image categorization and retrieval~\cite{silla2011survey}. This is done by integrating  hierarchy obtained from the biological taxonomy or from lexical databases to learn feature representations that capture semantic similarity between the categories. \cite{brust2020integrating} uses WordNet to construct hierarchies and incorporates the domain knowledge obtained from it into a \gls{cnn} classifier to improve the network's performance. \cite{bertinetto2020making} extends the traditional cross-entropy loss used in flat classification to work on hierarchies. To do so, they compute the conditional probability of the nodes at each layer of the hierarchy and calculate the cross-entropy loss at each layer. 
DeViSE~\cite{frome2013devise} transfers semantic knowledge from text to visual object recognition. It involves pre-training a language model to learn word embeddings and re-training a deep visual model using these embeddings for image label prediction.
\cite{barz2019hierarchy} uses an embedding algorithm that maps images onto a hypersphere. The distances on this hypersphere represent similarities that are derived from the lowest common ancestor height in a given hierarchy tree. One limitation of these methods is their reliance on an existing domain knowledge, which can be difficult to obtain or irrelevant in some contexts. 

A line of work relies on the visual features learnt by the network. \cite{yang2018retrieving} uses metric loss while training the network for image retrieval. Metric loss tries to pull closer the representations of images belonging to the same class, while pushing apart the inter-class representations. While this reinforces the condition that the intra-class representations are clustered tightly, it does not impact the semantic relationship between the instances. \cite{zhang2018image} constructs a visual hierarchy from the distance values between the image features and uses it to improve classification.
Our method is similar to \cite{park2002ranking}, which uses Ward clustering for constructing the hierarchy and uses it for re-ranking the images retrieved using a distance metric. A drawback of using Ward clustering is that they are sensitive to outliers and can sometimes group classes with low semantic similarities. In contrast, our approach groups classes only when their features overlap, which results in a more meaningful and semantically relevant hierarchy.

\section{Method}
This section explains the two main steps of our method: the construction of the hierarchy from the learned feature representations, and the calculation of the distance metric for ranking the images. 

\paragraph{Notations:} Consider a \gls{cnn} classifier trained on $K$ classes, each class represented as $\{C_1,..., C_K\}$. The feature vectors are extracted from the penultimate layer of the \gls{cnn}. We apply dimensionality reduction on these feature vectors using PCA to reduce the redundant features and speed up the computation~\cite{wu2000dimensionality}. We automatically determine the number of principal components required to explain 95\% of the original data variance by examining the eigenvalues in decreasing order.
Let the feature vectors corresponding to the training images be $x_{train}$. Assume that the class features follow multivariate Gaussian distributions with mean $\mu_c$ and covariance $\boldsymbol{\Sigma}_c$, where $c=\{1,..,K\}$. Note that this assumption can be broken when the intra-class variance is large. In such cases, one can constrain the latent space to be Gaussian using the methodology in \cite{venkataramanan2023self}. 
Given a query image $q$, its feature vector is denoted as $x_q$. Suppose the database consists of $N$ images, their corresponding feature vectors are denoted as $\{x_1,..,x_N\}$. 

\subsection{Hierarchy Construction}

The hierarchy construction involves identifying the classes with overlapping distributions of feature vectors in $x_{train}$. Inter-class overlapping typically occurs when there are visual and semantic similarities in the class images. To construct the hierarchy, a bottom-up approach is used, where the overlapping classes in the latent space are merged in the higher hierarchy level. The overlapping classes are identified using the \gls{bc}, which is a statistical metric to measure the amount of overlap between two statistical populations.

The Bhattacharya distance between two classes $C_i$ and $C_j$ is given by:
\begin{equation}
    D_B(C_i, C_j) = \frac{1}{8} (\mu_i - \mu_j)^T \boldsymbol{\Sigma}^{-1}(\mu_i - \mu_j) + \frac{1}{2} \ln \left(\frac{\det \boldsymbol{\Sigma}}{\sqrt{ \det \boldsymbol{\Sigma_i} \det \boldsymbol{\Sigma_j} }} \right)
\end{equation}
where $\boldsymbol{\Sigma} = \frac{\boldsymbol{\Sigma_i} + \boldsymbol{\Sigma_j}}{2}$

The \gls{bc} is calculated as:
\begin{equation}
    BC(C_i, C_j) = e^{-D_B(C_i, C_j)}
\end{equation}

The value of \gls{bc} ranges from 0 to 1 inclusive, where a value of 0 signifies no overlap and 1 means full overlap. If the value of the \gls{bc} between two classes is greater than or equal to a specified threshold $t$, then they are considered to be overlapping. Every cluster of classes that overlap with each other are identified. All classes within each cluster are merged into a unified class at a higher level of the hierarchy.
On the other hand, if the \gls{bc} is less than $t$, then these classes are considered to be distinct and are kept separate in the higher hierarchy level.
The $t$ value is a hyperparameter that depends on the dataset. The process of \gls{bc} estimation is repeated at each level on the new set of classes, and the overlapping classes are merged to create a hierarchy structure. This is performed until there are no overlaps between the classes, or all the classes have been merged. At the higher hierarchy levels, some classes include feature vectors from multiple small classes that were merged. At every level, the $\boldsymbol{\mu}$ and $\boldsymbol{\Sigma}$ values are recalculated from the feature vectors of the merged classes. 

\subsection{Distance Calculation} \label{sec:distance}
After constructing the hierarchy, we use it to perform the similarity search for image retrieval. The distances are calculated between the query feature $x_q$ and the image features $x_i$ in the database. Our final distance metric for ranking the images is a weighted sum of two distance values: cosine distance and a hierarchy based distance. 

\paragraph{Cosine distance:} The cosine distance is a similarity measure between two feature vectors and is popularly used in image retrieval tasks. The cosine distance between $x_q$ and $x_i$ is given by:
\begin{equation}
    D_C (x_q, x_i) = 1 - \frac{x_q \cdot x_i}{\lVert x_q \rVert \lVert x_i \rVert}
\end{equation}
where $\cdot$ is the dot product and $\lVert ~ \rVert$ is the L2 norm.

\paragraph{Hierarchical distance:} The hierarchy is a tree structure $\mathcal{H} = (V,E)$  where $V$ represents the nodes and $E$, the edges. $\mathcal{H}$ consists of $K$ leaf nodes. We want to identify the position of any database image in the hierarchy, which means finding the leaf closest to each image. To identify the leaf nodes, we calculate the \gls{md} between the feature vectors $x_i$ and the feature vectors $x_{train}$ corresponding to each leaf node. The \gls{md} between $x_i$ and node $c$, is given as:
\begin{equation}\label{eq:md}
    MD_c(x_i)= \sqrt{(x_i-\mu_c)^T\boldsymbol{\Sigma_c}^{-1}(x_i-\mu_c)}
\end{equation}
$x_i$ is assigned to the leaf node $n_i$ corresponding to the smallest \gls{md} value. 

Similarly, the leaf node $n_q$ corresponding to $x_q$ is identified. The distance calculation uses the height of the lowest common ancestor (LCA), where the LCA of two nodes is the parent node that is an ancestor of both nodes and does not have any child that is also an ancestor of both nodes. The hierarchical distance~\cite{barz2019hierarchy} between $x_q$ and $x_i$ is calculated as:
\begin{equation}
    D_H(x_q, x_i) = \frac{\mathrm{height(LCA(n_i,n_q))}}{\mathrm{height} (\mathcal{H})}
\end{equation}


The final distance is given as:
\begin{equation}
    D = D_C(x_q, x_i) + \alpha D_H(x_q, x_i)
\end{equation}
where $\alpha$ is a hyperparameter. 

The images are sorted in ascending order of $D$ to rank the images based on similarity and retrieve them.

\subsection{Robustness Analysis}
Robustness analysis demonstrates the ability of the image retrieval methods to maintain accurate and reliable performance even when the images are distorted or corrupted. The distortions introduced during the robustness analysis mimic real-world scenarios where images can be affected by factors like  blurring, occlusions, and changes in lighting conditions. 

We consider the following scenarios: (i) Blurring – The images are subject to Gaussian blurring, simulating the effect of an out-of-focus camera or a low-quality image (ii) Saturation change – The saturation values of the original images are modified to simulate variations in colour intensity. (iii)  Occlusions – Random occlusions are introduced in different parts of the images, simulating objects or regions being partially or completely blocked. 

\section{Experiments}

\subsection{Baselines}
We compare our method against the following approaches: (i) Image retrieval using Euclidean Distance, (ii) Image retrieval using Cosine Distance, (iii) Our method using hierarchical distance only, (iv) Semantic Embeddings~\cite{barz2019hierarchy}, (v) Our method applied on the features extracted from semantic embeddings in (iv), (vi) Distance calculation using the method in Sec.~\ref{sec:distance} on hierarchy constructed using Ward clustering~\cite{park2002ranking}, (vii) Distance calculation on semantic hierarchy obtained using domain knowledge.
\subsection{Experimental Setup}
We perform our experiments on three datasets with fine-grained visual features:

\textbf{CUB-200-2011~\cite{wah2011caltech}. } This is a widely used dataset for image retrieval, consisting of 200 classes of birds and a total of 11,788 images. The train data consists of 5,994 images and the test data consists of 5,794 images. The images were resized to $224\times224$, and random horizontal flipping and rotation ($\pm 30^{\circ}$) were used for data augmentation. Following~\cite{barz2019hierarchy}, we used ResNet-50~\cite{he2016deep} for training with an SGD optimizer. The network was trained for 100 epochs with a batch size of 12. The values of $t=0.30$ and $\alpha=3$ are obtained using hyperparameter search.
    
\textbf{CIFAR100~\cite{krizhevsky2009learning}. } It consists of 60,000 images, of size $32\times32$ belonging to 100 classes of wide range of objects. The training dataset consists of 50,000 images and the remaining are test data.  Following~\cite{barz2019hierarchy}, we used ResNet-110~\cite{he2016deep} for training with an SGD optimizer. The network was trained for 400 epochs with a batch size of 24. The values of $t=0.20$ and $\alpha=5$ are obtained using hyperparameter search.
    
\textbf{Diatom Dataset.} The diatom dataset consists of 9895 individual diatom images belonging to 166 classes~\cite{venkataramanan2023usefulness}. 
The network was trained using EfficientNet~\cite{tan2019efficientnet}. An SGD optimizer with a learning rate of 0.0002 was used. The images were resized to $256\times256$. This dataset contains high intra-class variance with distinct feature representations per class, and hence the latent space is not Gaussian, making it incompatible for \gls{bc} calculation. The network was trained using the method in \cite{venkataramanan2023self}, which ensures a Gaussian latent space. The training was performed for 100 epochs with a batch size of 12.
The values of $t=0.25$ and $\alpha=3$ are obtained using hyperparameter search.

For the CIFAR100 and CUB-200-2011, 80\% of the train data was used for training and 20\% for validation. For the diatoms, 70\% of the dataset was used for training, 10\% for validation and 20\% for testing. The networks were trained on GeForce RTX 3090 with 24 Gb of RAM.
The semantic hierarchy of CUB-200-2011 and CIFAR100 for the baseline experiments were obtained from \cite{barz2019hierarchy}. The diatom hierarchy was obtained from the taxonomy of the species, consisting of two levels: 'genus' and 'species'. The same model architectures were used for comparing all the baselines.

\textbf{Setup for Robustness Analysis. } The analysis is performed on the CUB-200-2011 dataset with the following settings. (i) A Gaussian Blur with kernel size (11,11) was applied to all the images, (ii) The saturation of images were adjusted with randomly chosen values in the range [0,1,0.5,1,2,5] and (iii) A random crop of dimension (70,70) was replaced by black patch in every image. The networks were retrained on the modified images, with the same training parameters. 

\subsection{Evaluation Metrics}
We use MAP@k values for evaluating the image retrieval performance. MAP@k measures the average precision (AP) of the top k retrieved images, and then takes the mean across all queries. A higher value indicates better performance.

\section{Results}

\subsection{Retrieval Results}

\begin{figure}[t]
    \centering
    \includegraphics[width=0.3\linewidth]{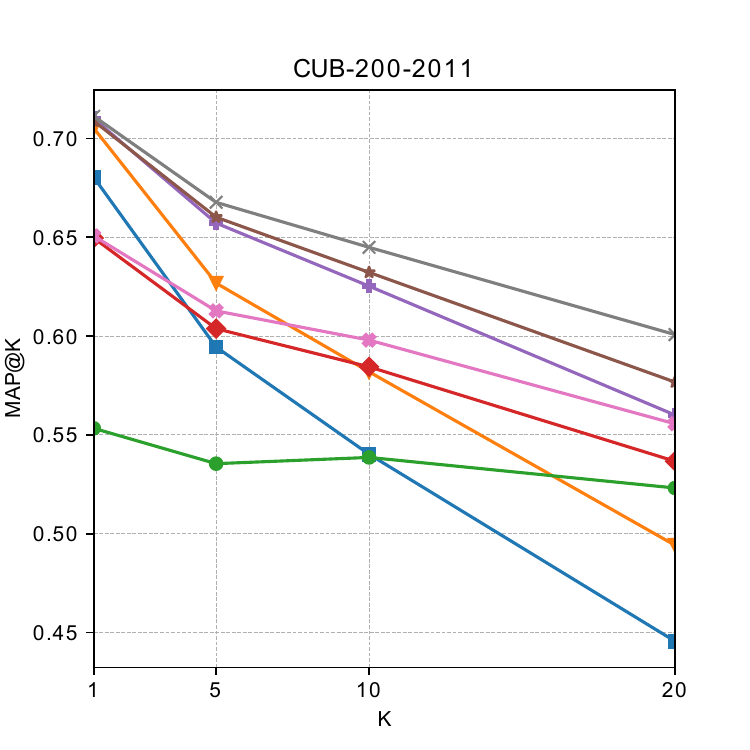}
    \includegraphics[width=0.3\linewidth]{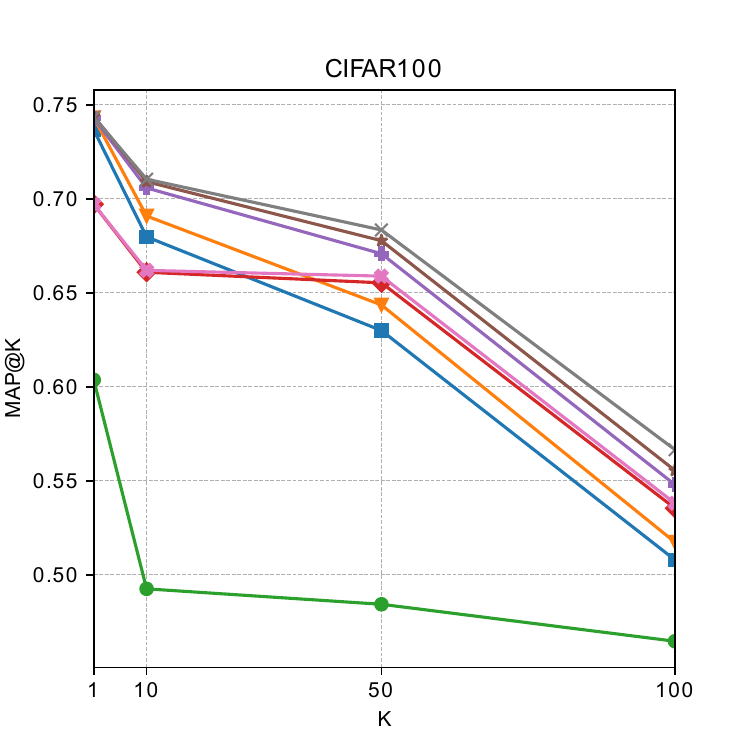}
    \includegraphics[width=0.3\linewidth]{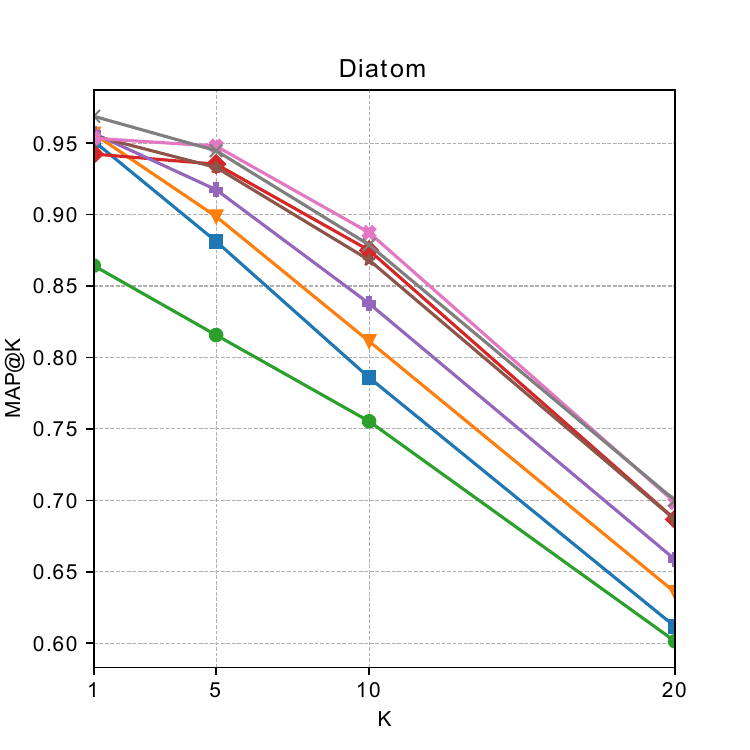}\\
    \includegraphics[width=\linewidth]{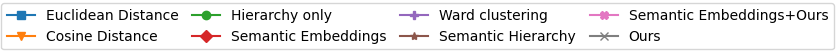}
    \caption{Retrieval performance on CUB-200-2011, CIFAR100 and Diatom dataset. Our method achieves state-of-the-art performance. The results suggest that incorporating hierarchy into content-based image retrieval is effective in improving the retrieval performance.}
    \label{fig:map_plots}
\end{figure}

Fig.~\ref{fig:map_plots} shows the plots of mAP@k values for different values of k for the three datasets. The results suggest that combining the visual and semantic cues improves the retrieval. Overall, our method achieves the best performance over the other baselines. 

\paragraph{Effect of distances:} The cosine distance retrieves images that are visually similar to the query, whereas, the hierarchical distance takes into account the semantic relationships between classes. Both these aspects are important when retrieving the images, and removing one of them deteriorates the performance. There is a significant drop in the MAP values in CIFAR100 when using only the hierarchical distance. One explanation for this is that CIFAR100 contains classes that are visually diverse, and hence, visual cues are more important over the semantic ones here. This could also explain the marginal improvement in the metrics when combining both the visual and semantic attributes than when using only the visual information for CIFAR100 over the other datasets.

\paragraph{Effect of hierarchy: }In the hierarchy-based methods, our approach outperforms the baselines. This is closely followed by our approach of distance calculation using semantic hierarchy. While semantic embeddings~\cite{barz2019hierarchy} also relies on visual and semantic features for the retrieval, its drop in performance could be attributed to the errors introduced due to disagreements between the visual and semantic relationships while training~\cite{brust2019not}. 
Our approach of hierarchy construction merges the classes only when there is a correlation between visual and semantic attributes, and thus it is robust against these errors. The Ward hierarchy approach may sometimes group classes that are semantically unrelated, resulting in less effective performance. 

\paragraph{Qualitative Analysis: } Fig.~\ref{fig:retrievals} shows some images retrieved using cosine distance and our method. While cosine distance retrieves images that are visually similar, some of them are semantically very different. Whereas, our method retrieves images that are visually and semantically meaningful.

\begin{figure}[t]
    \centering
    \includegraphics[width=0.3\linewidth]{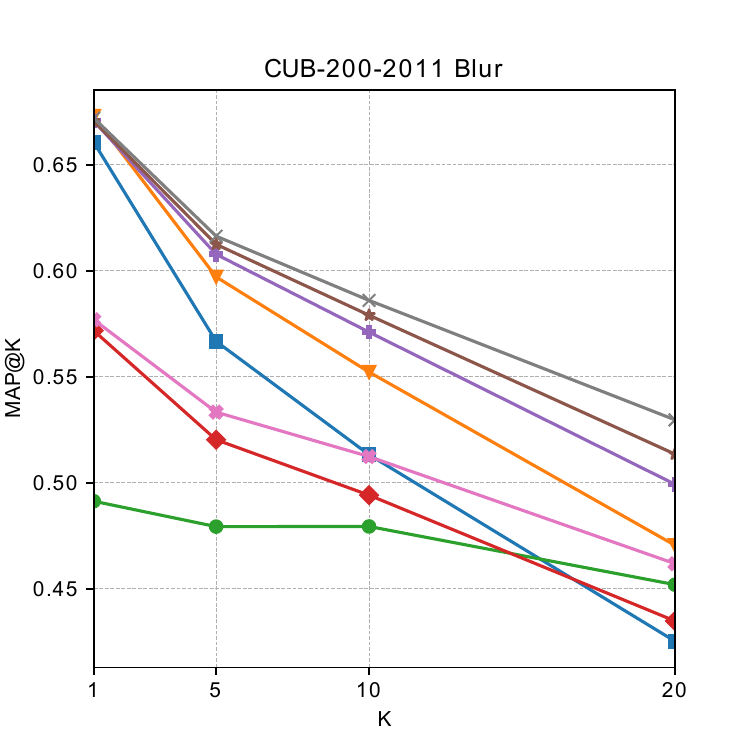}
    \includegraphics[width=0.3\linewidth]{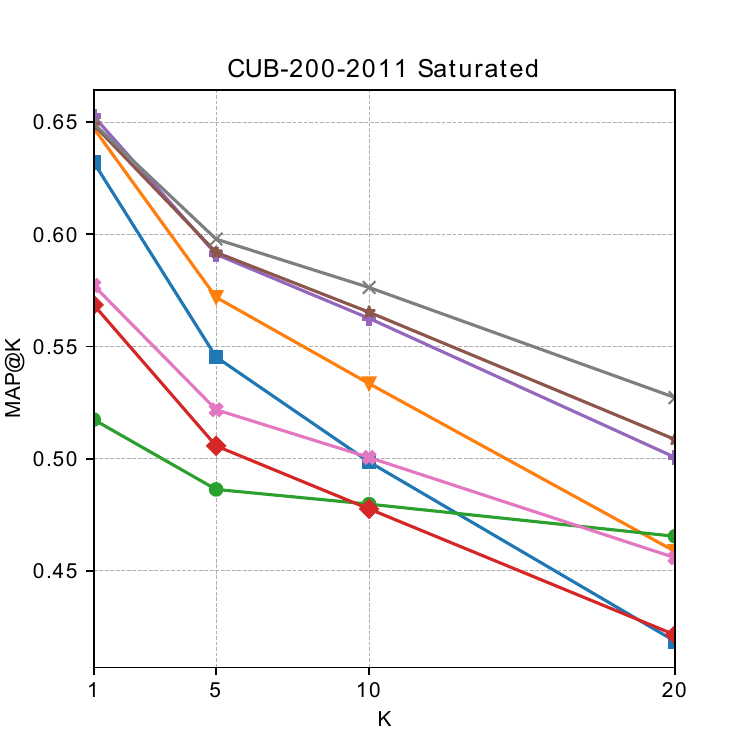}
    \includegraphics[width=0.3\linewidth]{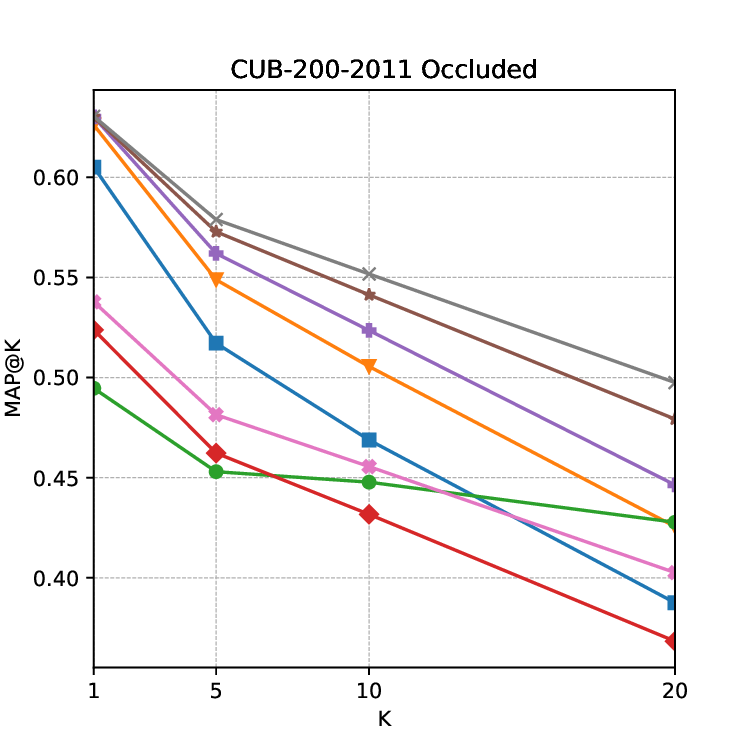}\\
    \includegraphics[width=\linewidth]{images/legend1.png}
    \caption{Robustness analysis on the CUB-200-2011 dataset. Blurring, saturation change and occlusion were applied to test the impact on the retrieval. Results show that our method still achieves the best performance.}
    \label{fig:robustness_map_plots}
\end{figure}

\paragraph{Robustness Analysis: } 
Fig.~\ref{fig:robustness_map_plots} shows the MAP@K plots for the robustness analysis on CUB-200-2011. Compared to the plot in Fig.~\ref{fig:map_plots}, one can observe a drop in the metrics. However, our method still achieves the best performance over the other baselines, indicating its robustness even in challenging conditions. 


\section{Discussion and Conclusion}
In this paper, we presented a method to tackle the problem of semantically dissimilar images retrieved in \gls{cbir}. By leveraging the learned feature representations from a CNN classifier, we construct a meaningful hierarchy that captures both visual and semantic information. This hierarchy is then integrated into the distance metric for similarity search, resulting in superior image retrieval performance, as demonstrated on fine-grained visual datasets. One limitation of our method arises when classes exhibit high variance in their features, resulting in potential overlaps with several or all other classes. Consequently, the network may incorrectly group them all into a single class when constructing the hierarchy. To mitigate this, we only group classes when the degree of overlap is above a threshold. One could also train the network using a metric loss to alleviate this, which we leave for future work. 
%
%
%
%

\bibliographystyle{splncs04}
\bibliography{biblio}

\end{document}